\documentclass[10pt,twocolumn,letterpaper]{article}

\usepackage{cvpr}
\usepackage{times}
\usepackage{epsfig}
\usepackage{graphicx}
\usepackage{amsmath}
\usepackage{amssymb}
\usepackage{array}

\usepackage{epstopdf}
\usepackage{multirow}
\usepackage{subfig}
\usepackage{amsmath}
\usepackage{amsthm}
\usepackage{amssymb}
\usepackage{dsfont}
\usepackage{algorithm}
\usepackage{algorithmic}
\usepackage{url}
\usepackage{cite}
\usepackage{tabularx}
\usepackage{booktabs}
\usepackage{enumitem}
\usepackage{url}
\usepackage{authblk}

\usepackage[pagebackref]{hyperref}
\hypersetup{
    bookmarks=false,
    bookmarksopen=false,
    bookmarksnumbered=true,
    pdfpagemode=UseOutlines,  
    colorlinks=true,          
    linkcolor=magenta,        
    anchorcolor=black,        
    citecolor=blue,           
    filecolor=black,          
    menucolor=black,          
    urlcolor=blue,            
    breaklinks=true,          
    pdfstartview={FitBH},
    pdfview={FitBH},
    pageanchor=false,
    pdftitle={Privacy Preserving Image-Based Localization},
    pdfauthor={Pablo Speciale, Johannes L. Sch{\"o}nberger, Sing Bing Kang, Sudipta N. Sinha, Marc Pollefeys}
}

\usepackage{fancyheadings}

\cvprfinalcopy 

\def\cvprPaperID{2426} 

\ifcvprfinal\fi

\usepackage{array,url,kantlipsum}

\usepackage[hang,flushmargin]{footmisc}

\DeclareMathOperator*{\argmin}{argmin}

\newcolumntype{Y}{>{\centering\arraybackslash}X}
\newcolumntype{R}{>{\raggedleft\arraybackslash}X}
\newcolumntype{L}{>{\raggedright\arraybackslash}X}

\renewcommand{\eqref}[1]{\hyperref[#1]{(}\ref{#1}\hyperref[#1]{)}}

\newcommand\customparagraph[1]{\vspace{0.3em}\noindent\textbf{#1}}

\usepackage[font=small,labelfont=bf]{caption}

\begin{document}

%

\title{\vspace*{-3.5ex} Privacy Preserving Image-Based Localization \vspace*{-3ex}}


%
%
\author{Pablo Speciale$^{1,2}$ \,\, Johannes L.\,Sch{\"o}nberger$^{2}$ \,\, Sing Bing Kang$^{2}$ \,\, Sudipta N.\,Sinha$^{2}$ \,\, Marc Pollefeys$^{1,2}$\\\vspace{-2pt}
$^1$ ETH Z{\"u}rich \qquad $^2$ Microsoft \vspace*{1.5ex}
}


\twocolumn[{%
\renewcommand\twocolumn[1][]{#1}%
\maketitle
\vspace*{-7ex}
\begin{center}
\vspace{-15pt}
\includegraphics[width=\linewidth]{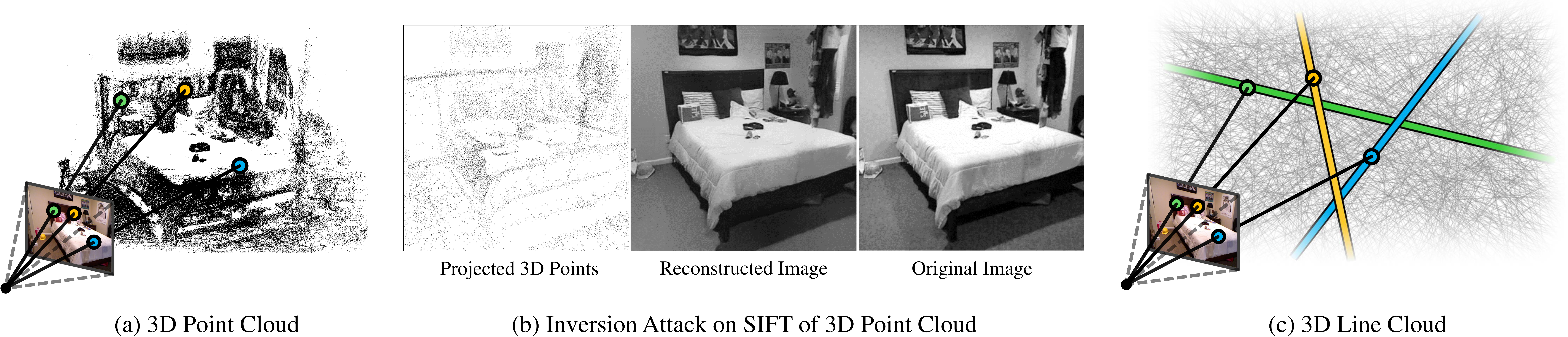}
\captionof{figure}{
\textbf{(a)} Traditional image-based localization using 3D point cloud, which reveals potentially confidential information in the scene.
\textbf{(b)} Reconstructed image using projected sparse 3D points and their SIFT features \cite{pittaluga2019}.
\textbf{(c)} Our proposed 3D line cloud protects user privacy by concealing the scene geometry and preventing inversion attacks, while still enabling accurate and efficient localization.}
\label{fig:teaser}
\end{center}

}]

\thispagestyle{empty}

\begin{abstract}
\vspace*{-1ex}
Image-based localization is a core component of many augmented/mixed reality (AR/MR) and autonomous robotic systems.
Current localization systems rely on the persistent storage of 3D point clouds of the scene to enable camera pose estimation, but such data reveals potentially sensitive scene information.
This gives rise to significant privacy risks, especially as for many applications 3D mapping is a background process that the user might not be fully aware of.
We pose the following question: How can we avoid disclosing confidential information about the captured 3D scene, and yet allow reliable camera pose estimation?
This paper proposes the first solution to what we call privacy preserving image-based localization.
The key idea of our approach is to lift the map representation from a 3D point cloud to a 3D line cloud.
This novel representation obfuscates the underlying scene geometry while providing sufficient geometric constraints to enable robust and accurate 6-DOF camera pose estimation.
Extensive experiments on several datasets and localization scenarios underline the high practical relevance of our proposed approach.
\end{abstract}


\vspace{-1ex}
\section{Introduction}
\label{sec:introduction}

Localizing a device within a scene by computing the camera pose from an image is a fundamental problem in computer vision, with high relevance in applications such as robotics \cite{Dube2017ICRA,Cummins2008IJRR,Schreiber2013IV}, augmented/mixed reality (AR/MR) \cite{Klein2009ISMAR,Lynen2015RSS}, and structure from motion (SfM) \cite{frahm2010building,schoenberger2015detail,schoenberger2016sfm}.
Arguably, the most common approach to image-based localization is structure-based \cite{irschara2009,Lynen2015RSS,Dube2017ICRA,Sattler2017PAMI} and tackles the problem by first matching the local 2D features of an image to the 3D point cloud model of the scene.
Geometric constraints derived from the matched 2D--3D point correspondences are then used to estimate the camera pose.
Inherently, the traditional approach to image-based localization thus requires the persistent storage of 3D point clouds.


The popularity of AR platforms such as ARCore~\cite{ARCore} and ARKit~\cite{ARKit}, wearable AR devices such as Microsoft HoloLens~\cite{Hololens}, and announcements of Microsoft's Azure Spatial Anchors (ASA)~\cite{asa}, Google's Visual Positioning System (VPS)~\cite{VPS} as well as 6D.AI's mapping platform~\cite{6Dai} indicate rising demand for image-based localization services that enable spatial persistence in AR/MR and robotics.
Even today, devices like HoloLens, MagicLeap1, or iRobot Roomba continuously map their 3D environment to operate.
This is a background process that users often are not consciously aware of.
As robotics and AR/MR become increasingly relevant in consumer and enterprise applications, more and more 3D maps of our environment will be stored on device or in the cloud and then shared with other clients.
Even though the source images are typically discarded after mapping, a person can easily
infer the scene layout and presence of potentially confidential objects based on a casual visual inspection of the 3D point cloud (see Fig.~\ref{fig:teaser}\hyperref[fig:teaser]{-(a)}).
Furthermore, methods that reconstruct images from local features (such as \cite{dosovitskiy2016,pittaluga2019}) make it possible to recover remarkably accurate images of the scene from point clouds (see Fig.~\ref{fig:teaser}\hyperref[fig:teaser]{-(b)}).
From a technical standpoint, these privacy risks have been widely ignored so far.
However, these will become increasingly relevant as localization services are adopted by more users as well as when mapping and localization capabilities will be more and more integrated with the cloud.
As a consequence, there has recently been a lot of discussion in the AR/MR community around the privacy implications of this development~\cite{privacy-manifesto-arcloud,ar-impact-internet,ar-thoughts}.


\begin{figure}[t]
	\includegraphics[width=\linewidth]{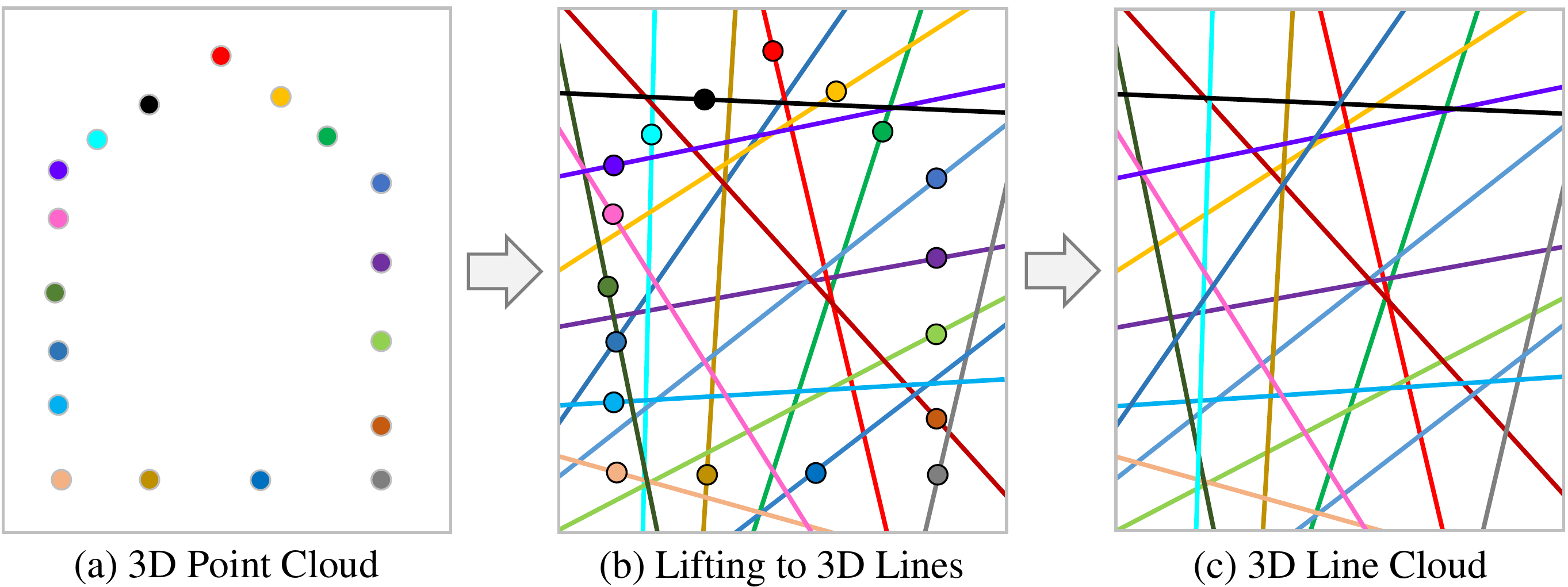}
	\caption{\textbf{3D Line Cloud.} The main idea of the proposed 3D line cloud representation that hides the geometry of the map.}
	\label{fig:p3p-p6l}
\end{figure}

In general, we predict three scenarios in which the privacy of users will be compromised.
First, if the scene itself is confidential (\eg, a worker in a factory or a person at home), then storing maps with a cloud-based localization service is inherently risky.
Running localization on trusted servers with maps stored securely could address privacy concerns, but even then the risk of unauthorized access remains.
In the second scenario, the scene itself is not confidential but there is a secret object or information (\eg, a hardware prototype in a workshop or some private details at home).
Yet, we still want to enable persistent localization in the same environment later on, without the risk of the secret information leaking via the 3D map of the scene.
This is especially relevant since users are typically not aware that mapping and localization services often continuously run in the background.
The final scenario involves low-latency and offline applications that need localization 
to run on client devices, which requires 3D maps to be shared among authorized users.
Obviously, distributing 3D maps with other users also compromises privacy.

To address these privacy concerns, we introduce a new research direction which we call {\em privacy preserving image-based localization} (see Fig.~\ref{fig:teaser}).
The goal is to encode the 3D map in a confidential manner (thus preventing sensitive information
from being
extracted), while maintaining the ability to perform robust and accurate camera pose estimation.
To the best of our knowledge, we are the first to propose a solution to this novel problem.

The key idea of our solution is to obfuscate the geometry of the scene in a novel map representation, where every 3D point is lifted to a 3D line with a random direction but passing through the original 3D point.
Only the 3D lines and the associated feature descriptors of the 3D points are stored, whereas the original 3D point locations are discarded.
We refer to such maps as \textit{3D line clouds} (see Fig.~\ref{fig:p3p-p6l}).
The 3D line cloud representation hides the underlying scene geometry and prevents the extraction of sensitive information.

%
%
%
%

To localize an image within a 3D line clouds, we leverage the traditional approach of feature matching \cite{irschara2009,Sattler2017PAMI} to obtain correspondences between local 2D image features and 3D features in the map.
Each correspondence provides the geometric constraint that the 2D image observation must lie on the image projection of its corresponding 3D line.
Based on this constraint, the problem of absolute camera pose estimation from 3D line clouds entails the intersection of a set of camera rays and their corresponding 3D lines in the map.
Towards leveraging this concept for privacy preserving localization, we show that a 3D line cloud can be interpreted as a generalized camera.
As a consequence, absolute camera pose estimation from 3D line clouds boils down to solving a generalized relative or absolute pose problem, which means we can repurpose existing algorithms~\cite{stewenius2005,kneip2013,li2008,Sweeney2015a,Sweeney2015,HeeLee2016} to solve our task.

In our paper, we study several variants of our
approach.
We first consider the case where the input is a single image and then generalize this concept to the case of jointly localizing multiple images.
We also present several specializations of our method for localization scenarios, where the
vertical direction or the scale of the scene is known.
These specializations are especially valuable in practical applications and underline the high relevance of our approach.

\customparagraph{Contributions.} We make the following contributions:
\textbf{(1)}~We introduce the \textit{privacy preserving image-based localization} problem and propose a first solution for it.
\textbf{(2)}~We propose a novel 3D map representation based on lifting 3D points to 3D lines, which preserves sufficient geometric constraints for pose estimation without revealing the 3D geometry of the mapped scene.
\textbf{(3)}~We propose minimal solvers for computing the camera pose given correspondences between 2D points in the image and 3D lines in the map. We study eight variants when the input is either a single image or multiple images, with and without the knowledge of the gravity direction or the scale of the scene.

\section{Related Work}
\label{sec:relatedwork}

\vspace{-1ex}
\customparagraph{Image-Based Localization.}
Recent progress in image-based localization has led to
methods that are now quite robust to changes in scene appearance
and illumination~\cite{arandjelovic2016,schonberger2018}, scale to large scenes~\cite{li2012worldwide,zeisl2015,sattler2015hyperpoints,Sattler2017PAMI}, and
are suitable for real-time computation
and mobile devices~\cite{arth2009,irschara2009,sattler2011,li2010,li2012worldwide,ventura2014b,lim2015,kendall2015} with compressed map representations~\cite{cao2014minimal,dymczyk2015}.
Traditional localization methods based on image retrieval~\cite{sivic2003video,jegou2012aggregating} and based on learning~\cite{kendall2015,weyand2016planet,brachmann2017dsac,walch2016image} have the advantage of not requiring the explicit storage of 3D maps.
However, model inversion techniques~\cite{mahendran2015understanding} pose privacy risks even for these methods.
Besides, they are generally not accurate enough~\cite{sattler2017large,walch2016image} to enable persistent AR and robotics applications.
Overall, to the best of our knowledge, there is no prior work on privacy preserving
image-based localization or on privacy-aware methods in other 3D vision tasks.

\customparagraph{Privacy-Aware Recognition.}
Privacy-aware object recognition and biometrics have been studied in vision since Avidan and Butman~\cite{avidan2006,avidan2007}, who
devised a secure face detection system. Other applications include
image retrieval~\cite{shashank2008}, face recognition~\cite{erkin2009}, video surveillance~\cite{upmanyu2009},
biometric verification~\cite{upmanyu2010}, activity recognition in videos to anonymize faces~\cite{ryoo2017,ren2018}, and detecting computer screens in first-person video~\cite{korayem2016}. A recent line of work explores learning data-driven models from private or encrypted datasets~\cite{cryptonets,yonetani2017,45428}.
All related works on privacy in computer vision focus on recognition problems, whereas ours is the first to focus on geometric vision.
While our work aims at keeping the scene geometry confidential, it is worth exploring confidential features as well. However, this is
beyond the scope of this paper.

\customparagraph{Privacy Preserving Databases.}
Privacy preserving techniques have been studied for querying data without leaking side information~\cite{Dinur2003}. Differential privacy~\cite{dwork2008} and k-anonymity~\cite{sweeney2002k} have been applied to the problem of location privacy~\cite{andres2013geo,ardagna2007,gedik2008}.
Learning data-driven models from private datasets has also received attention~\cite{cryptonets,yonetani2017,45428}.
However, existing techniques are inapplicable
for geometric vision problems such as image-based localization.

\customparagraph{Generalized Camera Pose Estimation.}
An important insight we present in the paper is that privacy preserving camera pose estimation from 3D line clouds has a close relation
to generalized cameras. After Grossberg and Nayar~\cite{grossberg2001} formulated the theory
of generalized cameras, Pless~\cite{pless2003} derived generalized epipolar constraints from the the Pl\"ucker representation of 3D lines.
Stewenius \etal~\cite{stewenius2005} proposed the first minimal solver for the generalized relative pose problem, whereas numerous
other solvers have been proposed for various generalized pose problems~\cite{Nister2007,ess2007,li2008,kneip2014a,kneip2014,thirthala2012,Lee2014,Sweeney2014,Sweeney2015,Sweeney2015a,camposeco2016,camposeco2018,ventura2014a}.

Generalized cameras are mostly used to model rigid multi-camera rigs or for dealing with multiple groups of calibrated cameras with known extrinsics~\cite{Sweeney2014,Sweeney2015,Sweeney2015a}. In those settings, generalized cameras typically have a small number of pinhole cameras with several observations per image. In contrast, our 3D line clouds can be viewed as generalized cameras with one pinhole camera (and one observation) per 3D line. While existing generalized pose solvers can be prone to degeneracies,
we avoid this problem by choosing lines with random directions. This not only enhances privacy but also leads to better conditioning of the problem. 
\vspace{-1ex}\section{Proposed Method}

In this section, we describe our proposed solution to privacy preserving image-based localization.
To give context, we first introduce the traditional approach to this problem for a single camera and then present the key concepts behind our privacy preserving method. We then describe
the extension of these concepts for jointly localizing multiple cameras. Finally, we discuss practical solutions
for several special cases, where the gravity direction is known or where we can obtain a local reconstruction
of the scene with known or unknown scale. In our description, we focus on the high level intuitions behind our approach and refer the reader to related literature
for details on the underlying algorithms needed to solve the various cases.

\subsection{Traditional Camera Pose Estimation}

\begin{figure}[t]
    \includegraphics[width=\linewidth]{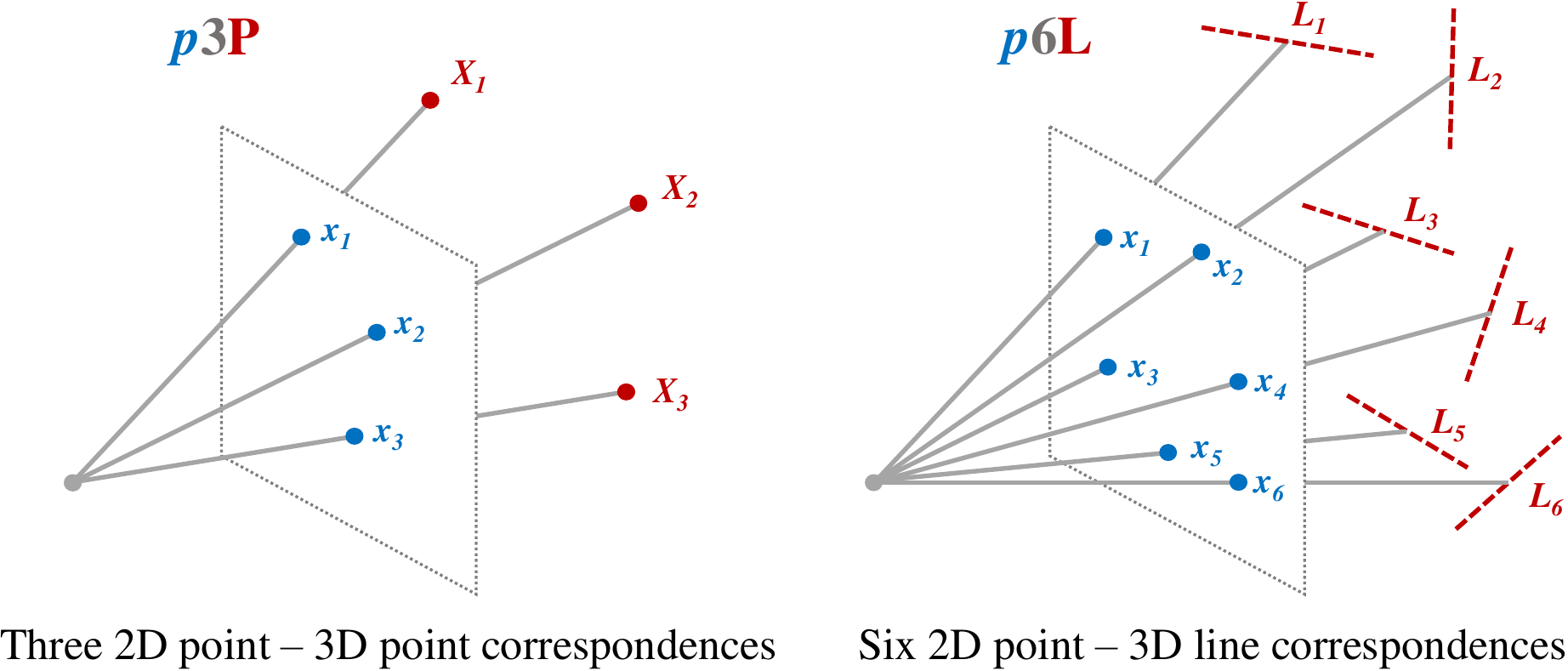}
    \caption{\textbf{Camera Pose Estimation.} Left: using traditional 3D point cloud. Right: using our privacy preserving 3D line cloud.}
    \label{poseEstSingle}
\end{figure}

We follow the traditional approach of structure-based visual localization \cite{irschara2009,Sattler2017PAMI}, where the map of a scene is represented by a 3D point cloud, which is typically reconstructed from images using SfM \cite{schoenberger2016sfm}.
To localize a pinhole camera with known intrinsics in the reconstructed scene, one estimates its absolute pose $\boldsymbol{P} = \begin{bmatrix} \boldsymbol{R} & \boldsymbol{T} \end{bmatrix}$ with $\boldsymbol{R} \in \text{SO(3)}$ and $\boldsymbol{T} \in \mathds{R}^3$
from correspondences between normalized 2D observations $\boldsymbol{x} \in \mathds{R}^2$ in the image and 3D points $\boldsymbol{X} \in \mathds{R}^3$ in the map.
To establish 2D--3D correspondences, the classical approach is to either use direct or indirect matching from 2D image features to 3D point features \cite{irschara2009,Sattler2017PAMI}.
Each 2D--3D point correspondence provides two geometric constraints for absolute camera pose estimation in the form of
\begin{equation}
\boldsymbol{0} = \boldsymbol{\bar{x}} - \boldsymbol{P} \boldsymbol{\bar{X}} = \lambda  \begin{bmatrix} \boldsymbol{x} \\ 1 \end{bmatrix} - \boldsymbol{P} \boldsymbol{\bar{X}} \enspace ,
\label{eq:point-to-point-constraint}
\end{equation}
where $\lambda$ is the depth of the image observation $\boldsymbol{x}$ while $\boldsymbol{\bar{x}} \in \mathds{P}^2$ and $\boldsymbol{\bar{X}} \in \mathds{P}^3$ are the lifted representations of $\boldsymbol{x}$ and $\boldsymbol{X}$ in projective space, respectively.
Naturally, we need a minimum of three 2D--3D correspondences to estimate the six unknowns in $\boldsymbol{P}$.
In the general case, this problem is typically referred to as the \textit{p}nP problem and in the minimal case as the \textit{p}3P problem.
Since the matching process is imperfect and leads to outliers in the set of 2D--3D correspondences, the standard procedure is to use robust algorithms such as RANSAC~\cite{fischler1981random,raguram2013usac} in combination with efficient minimal solvers to optimize Eq.~\eqref{eq:point-to-point-constraint} for computing an initial pose estimate.
Subsequently, that estimate is then refined by solving the non-linear least-squares problem
\begin{equation}
\boldsymbol{P}^* = \argmin_{\boldsymbol{P}} \Vert \boldsymbol{\bar{x}} - \boldsymbol{P} \boldsymbol{\bar{X}} \Vert_2 \enspace ,
\label{eq:point-to-point-refinement}
\end{equation}
which gives the maximum likelihood estimate based on a Gaussian error model $\boldsymbol{x} \sim \mathcal{N}(\boldsymbol{0}, \boldsymbol{\sigma}_{\boldsymbol{x}})$ for the image observations.
This approach has been widely used \cite{irschara2009,zeisl2015,lim2015,Lynen2015RSS,Sattler2017PAMI} and enables efficient and accurate image-based localization in large scenes.
However, it requires knowledge about the scene geometry in the form of the 3D point cloud $\boldsymbol{X}$ and thereby this approach
inherently reveals the geometry of the scene.
In the next sections, we present our novel localization approach that overcomes this privacy limitation.

\subsection{Privacy Preserving Camera Pose Estimation}

The core idea behind our approach to enable privacy preserving localization is to obfuscate the geometry of the map in a way that conceals information about the underlying scene without losing the ability to localize the camera within the scene.
In order to obfuscate the 3D point cloud geometry, we lift each 3D point $\boldsymbol{X}$ in the map to a 3D line~$\boldsymbol{L}$ with a random direction $\boldsymbol{v} \in \mathds{R}^3$ passing through $\boldsymbol{X}$.
The lifted 3D line $\boldsymbol{L}$ in Pl\"ucker coordinates \cite{pless2003} is defined as
\begin{equation}
\boldsymbol{L} = \begin{bmatrix} \boldsymbol{v} \\ \boldsymbol{w} \end{bmatrix} \in \mathds{P}^5 \qquad \text{with} \qquad \boldsymbol{w} = \boldsymbol{X} \times \boldsymbol{v} \enspace .
\label{eq:plucker-line}
\end{equation}
Importantly, since direction $\boldsymbol{v}$ is chosen at random and due to the cross product being a rank-deficient operation, the original 3D point location $\boldsymbol{X}$ cannot be recovered from its lifted 3D line $\boldsymbol{L}$.
We only know that $\boldsymbol{L}$ passes through $\boldsymbol{X}$ somewhere and that this also holds for their respective 2D projections $\boldsymbol{l}$ and $\boldsymbol{x}$ in the image.
Formally, a 2D image observation $\boldsymbol{x}$ passes through the projected 2D line $\boldsymbol{l}$, if it satisfies the geometric constraint
\begin{equation}
0 = \boldsymbol{\bar{x}}^T \boldsymbol{l}
\quad \text{with} \hspace*{1ex} [ \boldsymbol{l}]_{\times}
    = \left[ \begin{smallmatrix}
        0 & - l_3 & l_2 \\
        l_3 & 0 & - l_1 \\
        - l_2 & l_1 & 0 \\
      \end{smallmatrix} \right]
    = \boldsymbol{P} [\boldsymbol{L} ]_{\times}  \boldsymbol{P}^T,
\label{eq:point-to-line-constraint}
\end{equation}
where $[\boldsymbol{L} ]_{\times}$ is the Pl\"ucker matrix defined as
\begin{equation}
    [\boldsymbol{L} ]_{\times} = \begin{bmatrix}
                    - [\boldsymbol{w}]_{\times} & -\boldsymbol{v} \\
                    \boldsymbol{v}^T            &       0         \\
                  \end{bmatrix} \enspace .
\end{equation}
Using this constraint for absolute camera pose estimation requires a minimum of six 2D point to 3D line correspondences to solve for the six unknowns in $\boldsymbol{P}$.
This is in contrast to the traditional approach, where each correspondence provides two constraints and thus only three correspondences are needed.
Similar to the traditional \textit{p}nP and \textit{p}3P problems, we denote the general problem as {\textit{p}nL} and the minimal problem as {\textit{p}6L}.
Geometrically, solving the \textit{p}nL problem is equivalent to rotating and translating the bundle of rays defined by $\boldsymbol{x}$ and passing through the pinhole of the camera, such that the bundle of camera rays intersect with their corresponding 3D lines in the map (see Fig~\ref{poseEstSingle}).
Note, this is a specialization of the generalized relative pose problem \cite{stewenius2005}, where the rays in the first generalized camera represent the known 3D lines of the map, and the rays of the second generalized camera represent the 2D image observations of the pinhole camera that we want to localize.

We embed this concept into the traditional localization pipeline by robustly estimating an initial pose estimate using RANSAC with the minimal solver by Stewenius~\etal~\cite{stewenius2005} to solve Eq.~\eqref{eq:point-to-line-constraint}.
We then non-linearly refine the initial pose by minimizing the geometric distance between the observed 2D point and the projected 3D line as
\begin{equation}
\boldsymbol{P}^* = \argmin_{\boldsymbol{P}} \frac{\boldsymbol{\bar{x}}^T \boldsymbol{l}}{\sqrt{l_1^2 + l_2^2}} \enspace .
\label{eq:point-to-line-refinement}
\end{equation}
After deriving the theory for a single camera in this section, we next generalize our approach to the joint localization of multiple images and the special case with known vertical.

%
%
\begin{table*}[t]
	\small
    \vspace*{-2ex}
	\begin{tabularx}{1.0\textwidth}{cc RR RR}
		\toprule
		\textbf{Constraints} & \textbf{\textit{Query Type}} & \multicolumn{2}{c}{\textbf{\textsc{Point to Point}} (Traditional)} & \multicolumn{2}{c}{\textbf{\textsc{Point to Line}} (Privacy Preserving)}  \\
		\cmidrule(rr){1-1} \cmidrule(rl){2-2} \cmidrule(rl){3-4} \cmidrule(ll){5-6}
		\multirow{2}{*}{{2D -- 3D}} & \textit{Single-Image}                 & \textit{p}3P~\cite{haralick1994}          & \textit{p}2P+u~\cite{Sweeney2015}           & \textit{p}6L~\cite{stewenius2005}          & \textit{p}4L+u~\cite{Sweeney2015a}             \\
		& \textit{Multi-Image}                  & m-\textit{p}3P~\cite{HeeLee2016}           & m-\textit{p}2P+u~\cite{horn1987}             & m-\textit{p}6L~\cite{stewenius2005}         & m-\textit{p}4L+u~\cite{Sweeney2015a}            \\
		\midrule
		\multirow{2}{*}{{3D -- 3D}} & \multirow{2}{*}{\textit{Multi-Image}} & m-P3P+$\lambda$\hspace*{2ex}\cite{umeyama1991least}   & m-P2P+$\lambda$+u\hspace*{2ex}\cite{umeyama1991least}   & m-P4L+$\lambda$\hspace*{2ex}\cite{Sweeney2014} & m-P3L+$\lambda$+u\hspace*{2.1ex}\cite{camposeco2018} \\
		&                                       & m-P3P+$\lambda$+s~\cite{horn1987}             & m-P2P+$\lambda$+u+s~\cite{horn1987}             & m-P3L+$\lambda$+s~\cite{HeeLee2016}                      & m-P2L+$\lambda$+u+s~\cite{Sweeney2015} \\
		\bottomrule
	\end{tabularx}
    \caption{\textbf{Camera Pose Problems}. Traditional methods are \textbf{\textit{p}{\scriptsize*}P} (2D point to 3D point) and \textbf{P{\scriptsize*}P} (3D point to 3D point), whereas privacy preserving methods are \textbf{\textit{p}{\scriptsize*}L} (2D point to 3D line) and \textbf{P{\scriptsize*}L} (3D point to 3D line). The methods in the first row localize single images, whereas the rest jointly localizes multiple images (prefix \textit{m}). We have general solvers as well specialized ones for known vertical direction (suffix \textit{+u}). The bottom two rows exploit known 3D structure (suffix $+\lambda$ and suffix \textit{+s} for known scale) local to the camera to be localized.}
	\label{table:methods}
\end{table*}

\vspace*{-2ex}
\subsubsection{Generalization to Multiple Cameras}

\begin{figure}[t]
    \includegraphics[width=\linewidth]{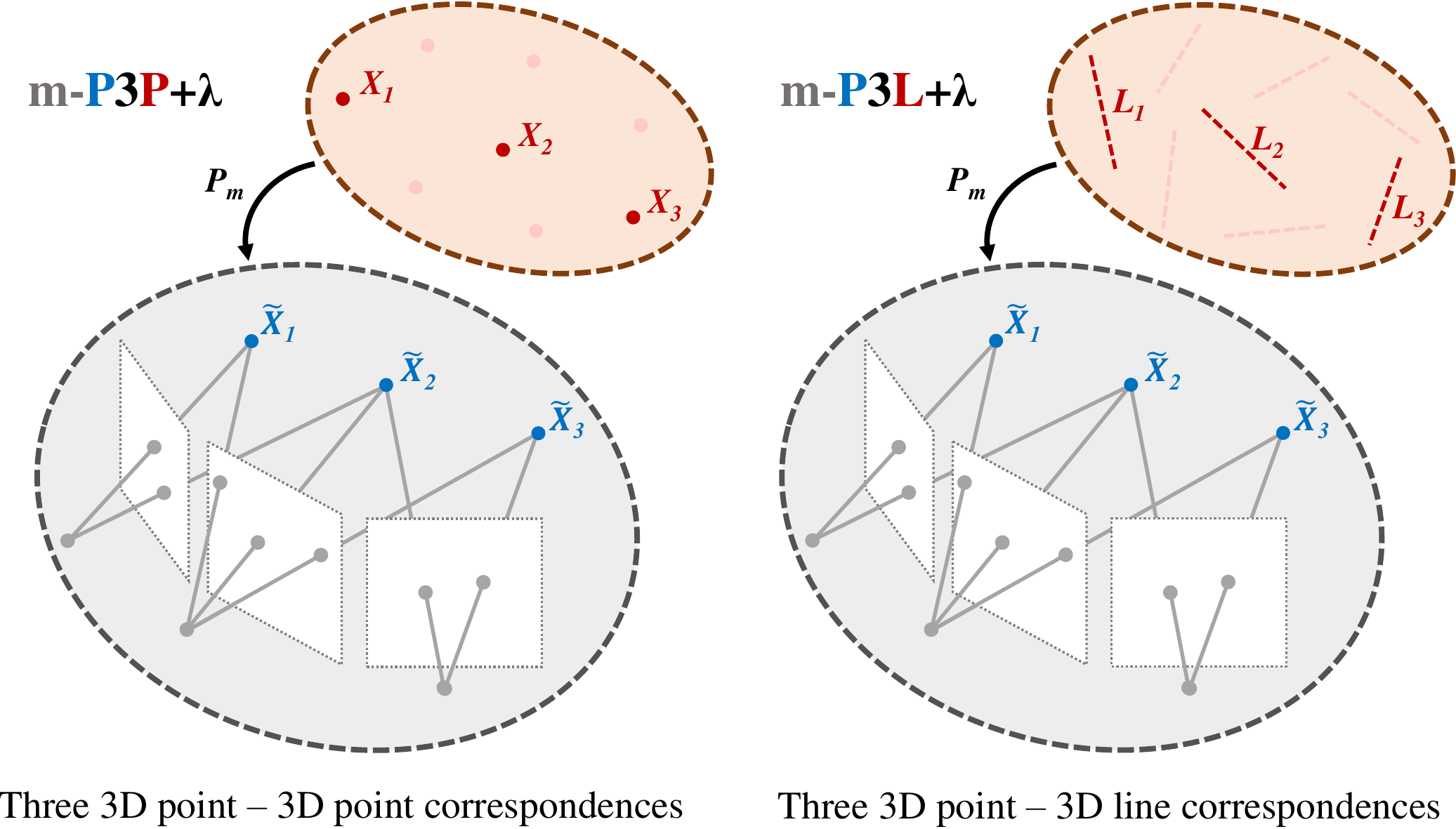}
    \caption{\textbf{Camera Pose Estimation with Known Structure.} Left: traditional setup with 3D point cloud maps. Right: our proposed approach using 3D line cloud maps.}
    \label{poseEstMult}
\end{figure}

While existing localization approaches typically only consider a single image \cite{irschara2009,zeisl2015,Lynen2015RSS,Sattler2017PAMI}, many devices such as head mounted displays, robots, or vehicles are equipped with multiple rigidly mounted cameras, that have been calibrated a priori.
Jointly localizing multiple cameras together brings great benefits for localization by leveraging the combined field of view to retrieve more 2D--3D correspondences and by reducing the number of unknown pose parameters for an increased redundancy in the estimation problem.
In addition, many mobile devices nowadays have built-in SLAM capabilities (\eg, ARKit, ARCore), which can be leveraged to take advantage of the same simplifications as with multi-camera systems by treating a local camera trajectory as an extrinsic calibration of multiple images.

The joint localization of multiple cameras differs from the case of a single camera primarily in how the problem is parameterized.
Instead of determining a separate pose $\boldsymbol{P} \in \text{SE}(3)$ for each camera, we reparameterize the pose as
\begin{equation}
\boldsymbol{P} = \boldsymbol{P}_c \boldsymbol{P}_m \quad \text{with} \quad \boldsymbol{P}_m = s_m \begin{bmatrix} \boldsymbol{R}_m & \boldsymbol{T}_m \\ \boldsymbol{0} & s_m^{-1} \end{bmatrix} .
\end{equation}
We now estimate only a single 3D similarity transformation $\boldsymbol{P}_m \in \text{Sim}(3)$, while the known relative extrinsic calibrations $\boldsymbol{P}_c$ of the individual cameras stay fixed.
Note that if we know the relative scale of $\boldsymbol{P}_c$ with respect to the 3D points $\boldsymbol{X}$ in the map, we can eliminate the scale factor $s_m \in \mathds{R}^+$ and reduce $\boldsymbol{P}_m$ to a 3D rigid transformation.

In the literature, this problem is referred as the generalized absolute pose problem \cite{Nister2007,HeeLee2016}, which is analogous to the traditional problem and does not conceal the 3D point cloud.
In most practical applications, we can assume that the scale $s_m = 1$, because multi-camera setups are typically calibrated to metric scale, and due to the fact that most SLAM systems recover scale from integrated inertial measurements.
In the following, we therefore initially restrict our work to the scenario where $\boldsymbol{P}_m \in \text{SE}(3)$.
We refer to the solution of this problem as \textit{m-\textit{p}nP} in the general case and \textit{m-\textit{p}3P} in the minimal case.
However, efficient solutions also exist for the more general case $\boldsymbol{P}_m \in \text{Sim}(3)$ \cite{ventura2014a,Sweeney2014}.

In the privacy preserving setting, the
generalization
to multiple images again
boils down to
solving a generalized relative pose problem~\cite{stewenius2005}.
However, the rays of the second generalized camera arise from 2D
image observations of multiple instead of a single pinhole camera.
We refer to the generalized solutions in the privacy preserving setting as \textit{m-\textit{p}}nL in the general and \textit{m-\textit{p}}6L in the minimal case.

\vspace*{-2ex}
\subsubsection{Pose Estimation with Known Structure}

So far, we have discussed a way to estimate the camera pose from the rays of 2D image observations directly. In many situations though, it is possible to obtain the depth $\lambda$ of an image observation $\boldsymbol{x}$, after which, its 3D location relative to the camera is computed as $\boldsymbol{\tilde{X}} = \lambda \boldsymbol{\bar{x}}$.
Such 3D data can be extracted through an active depth camera that yields RGB-D images or through multi-view triangulation.
In the traditional localization problem, we can therefore directly estimate the camera pose as the transformation that best aligns the two corresponding 3D point sets using the constraint
\begin{equation}
\boldsymbol{0} = \boldsymbol{\tilde{X}} - \boldsymbol{P} \boldsymbol{\bar{X}} \enspace .
\label{eq:3d-to-3d-constraint}
\end{equation}
To solve this equation in the minimal case, we need only three correspondences for the 6-DOF of the 3D rigid transformation $\boldsymbol{P}$.
Eq.~\eqref{eq:3d-to-3d-constraint} is typically solved in a least-squares fashion, and in this form has a direct and computationally efficient solution \cite{horn1987,umeyama1991least}; we refer to this as {m-PnP+$\lambda$} and {m-P3P+$\lambda$} in the general and minimal cases, respectively.

Similarly, we can also take advantage of the local 3D points $\boldsymbol{\tilde{X}}$ in our privacy preserving approach.
Instead of solving a generalized relative pose problem to find the intersection between the 3D lines of the map and the camera rays, we now try to find a pose such that the 3D lines $\boldsymbol{L}$ of the map pass through the 3D points $\boldsymbol{\tilde{X}}$.
We can formalize this in the following geometric constraint
\begin{equation}
\boldsymbol{0} = \boldsymbol{\tilde{X}} -  \boldsymbol{P} \begin{bmatrix} \boldsymbol{v} \times \boldsymbol{w} + \alpha \boldsymbol{v} \\ 1 \end{bmatrix} \enspace ,
\label{eq:3d-to-3d-line-constraint}
\end{equation}
where $\alpha$ is the unknown distance from the random origin $\boldsymbol{v} \times \boldsymbol{w}$ of the 3D line $\boldsymbol{L}$ to the secret 3D point $\boldsymbol{X}$.
By inverting the role of the camera and the map, this problem is geometrically equivalent to the generalized absolute pose problem, \ie, we can repurpose m-\textit{p}nP to solve for the unknown pose $\boldsymbol{P}$.
As such, we now only need a minimum of three 3D point to 3D line correspondences compared to the six correspondences needed to solve m-\textit{p}6L (see~Fig.\ref{poseEstMult}).
Note that requiring fewer points to solve the minimal problem is advantageous in RANSAC, which has an exponential runtime complexity in the number of sampled points.
The solution to Eq.~\eqref{eq:3d-to-3d-line-constraint} is also more efficient to compute \cite{HeeLee2016} as compared to m-\textit{p}6L.
We refer to this problem as {m-\textit{p}nL}$+\lambda$ in the general and {m-P3L}$+\lambda$ in the minimal case.

\begin{figure*}[t]
	\centering
    \vspace*{-2ex}
	\includegraphics[width=\textwidth]{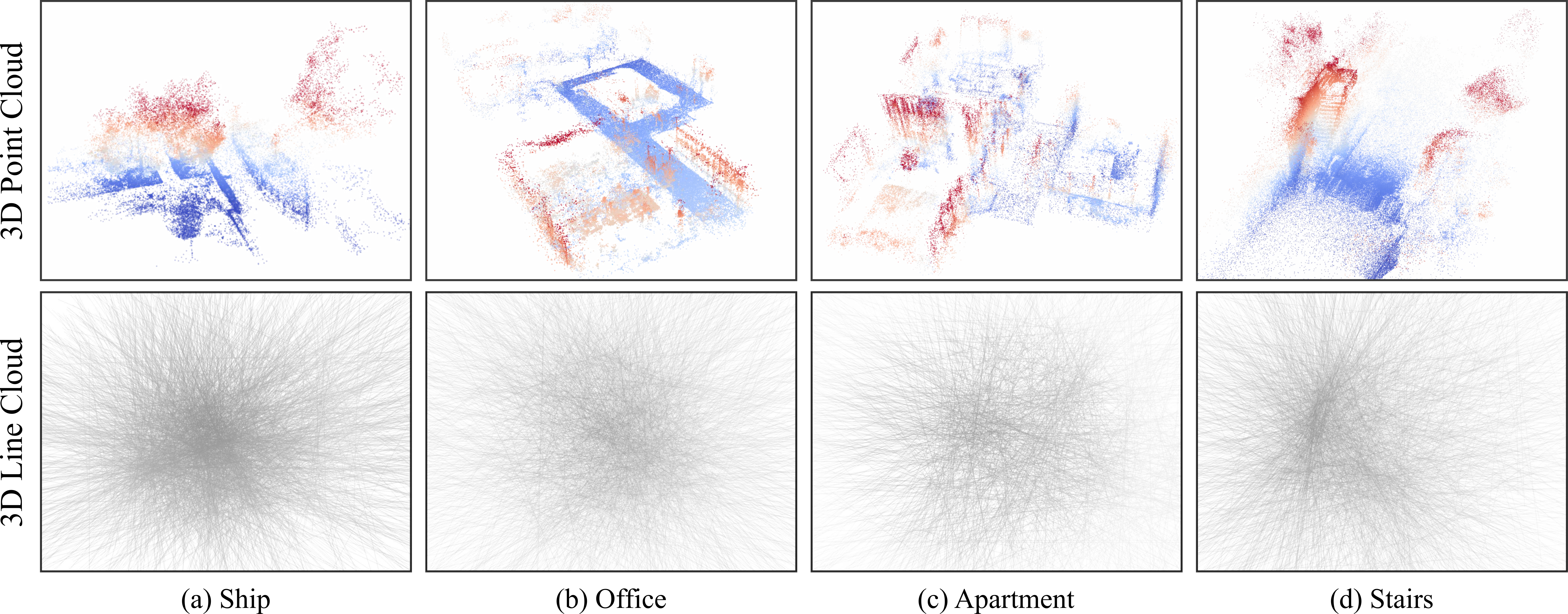}
	\caption{\textbf{Dataset Visualization.} The original 3D point cloud with the corresponding 3D line cloud is shown.}
	\label{fig:datasets}
\end{figure*}

\vspace*{-2ex}
\subsubsection{Extension to Unknown Scale}

The approach described in the previous section can be sensitive to inaccurate 3D point locations $\boldsymbol{X}$ and $\boldsymbol{\tilde{X}}$.
This is problematic, even if the two 3D point clouds have only slightly different scale, \eg, due to drift in SLAM or slight miscalibrations of the multi-camera system.
In comparison, the constraints used by \textit{p}nP and \textit{p}nL are less susceptible to this issue.
This is because the viewpoints used to triangulate $\boldsymbol{X}$ and $\boldsymbol{\tilde{X}}$ are inherently similar in image-based localization and the uncertainties $\sigma_\lambda$ in the depths $\lambda$ are usually larger than the uncertainties $\boldsymbol{\sigma}_{\boldsymbol{x}}$ in image space.

To overcome this issue, it is typically better to estimate a 3D similarity transformation $s \boldsymbol{P}$ with $s \in \mathds{R}^+$ instead of a 3D rigid transformation, when performing structure-based alignment.
The constraint in Eq.~\eqref{eq:3d-to-3d-constraint} then becomes
\begin{equation}
\boldsymbol{0} = \boldsymbol{\tilde{X}} - s \boldsymbol{P} \boldsymbol{\bar{X}} \enspace ,
\label{eq:3d-to-3d-constraint-with-scale}
\end{equation}
while the privacy preserving constraint in Eq.~\eqref{eq:3d-to-3d-line-constraint} becomes
\begin{equation}
\boldsymbol{0} = \boldsymbol{\tilde{X}} -  s \boldsymbol{P} \begin{bmatrix} \boldsymbol{v} \times \boldsymbol{w} + \alpha \boldsymbol{v} \\ 1 \end{bmatrix} \enspace .
\label{eq:3d-to-3d-line-constraint-with-scale}
\end{equation}
Now we need at least four correspondences to estimate the 7-DOF 3D similarity.
Note that Eq.~\eqref{eq:3d-to-3d-constraint-with-scale} has a comparatively simple and efficient solution \cite{umeyama1991least} that we refer to as {m-PnP+$\lambda$+s}.
In the privacy preserving setting, the problem of computing a 3D rigid transformation is exactly minimal, \ie, we now need a fourth correspondence to estimate the additional scale parameter using the constraints in Eq.~\eqref{eq:3d-to-3d-line-constraint-with-scale}.
This is equivalent to
the generalized absolute pose and scale problem \cite{Sweeney2014}, where the role of cameras and map is again reversed.
We refer to the reversed problem as m-PnL+$\lambda$+s in the general and as m-P4L+$\lambda$+s in the minimal case.

\subsubsection{Specialization with Known Vertical}

Oftentimes, an estimate of the gravity direction in both the reference frame of the camera and the 3D map may be available, \eg, from inertial measurements or vanishing point detection.
By pre-aligning the two reference frames to the vertical direction, we can reduce the number of rotational pose parameters from three to one such that $\boldsymbol{R} \in \text{SO}(2)$.
This parameterization of the rotation simplifies the geometric constraints, and leads to more efficient and numerically stable solutions for these problems.
In addition, 
the minimal cases require fewer points,
leading to a better runtime of RANSAC.
We implement the known gravity setting for all described problems and indicate this with the suffix $\textit{+u}$.
An overview of all the problems is given in Table~\ref{table:methods}.

\section{Experimental Evaluation}
\label{sec:results}

To demonstrate the high practical relevance of our approach, we conduct an extensive list of experiments on real-world data with ground-truth.
We evaluate the pose estimation performance in terms of accuracy/recall and robustness to the input by comparing our privacy preserving approach using 3D line clouds to the traditional approach of using 3D point clouds.
In the following, we first describe the experimental setup before presenting the results.

\subsection{Setup}

\begin{figure*}[t]
	\centering
    \vspace*{-2ex}
	\includegraphics[width=\textwidth]{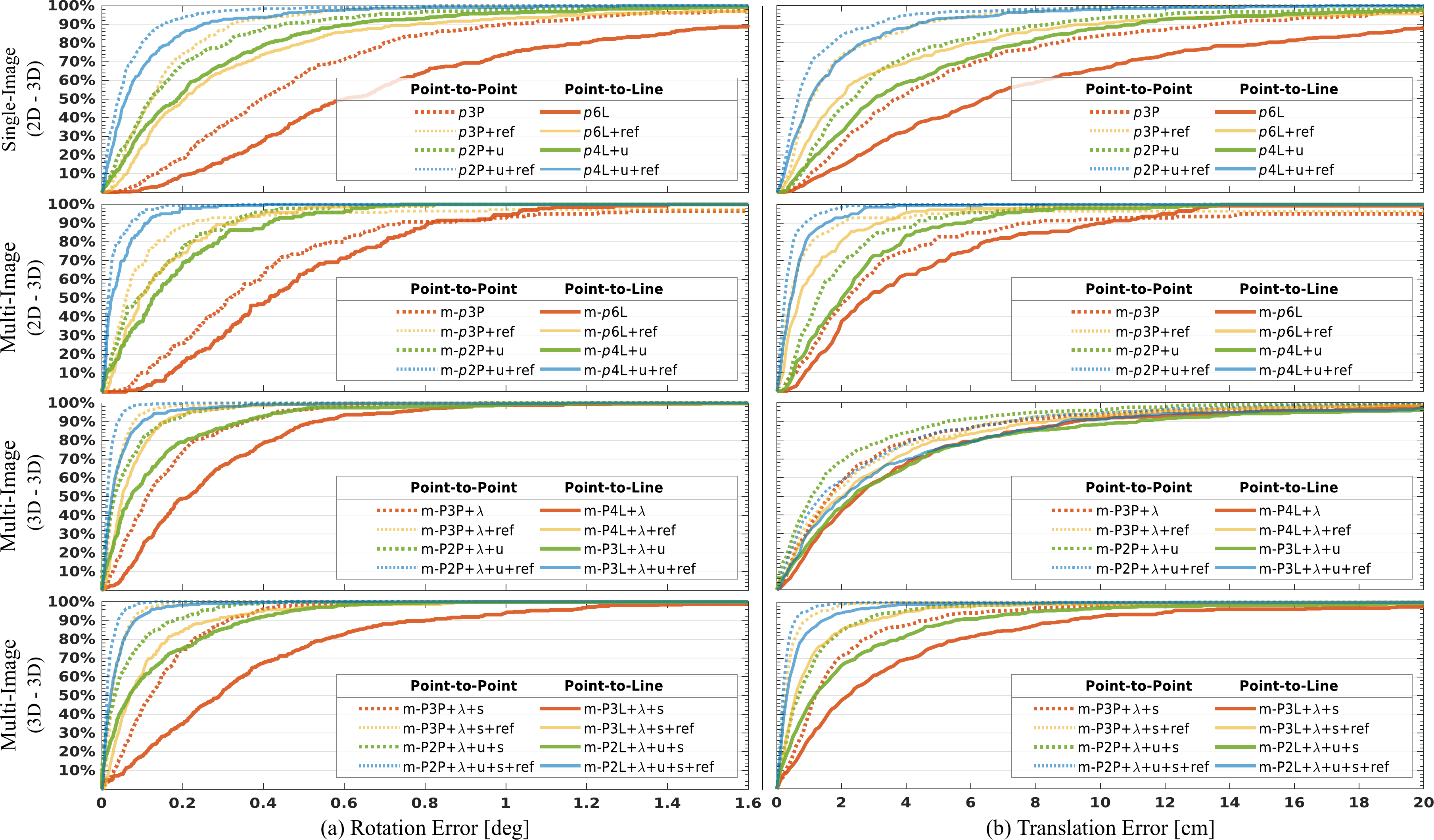}
	\caption{\textbf{Camera Pose Estimation Errors Plots.} Cumulative rotation and translation error histograms for all 16 evaluated methods.}
	\label{fig:errors_plots}
\end{figure*}

\vspace{-1ex}
\customparagraph{Datasets.}
We collect 15 real-world datasets of complex indoor and outdoor scenes (see Fig.~\ref{fig:datasets}) using a mix of mobile phones and the research mode of the Microsoft HoloLens~\cite{Hololens}.
To realistically simulate an image-based localization scenario, we captured \textit{map images} used to reconstruct a 3D point cloud of the scene and \textit{query images} from significantly different viewpoints used for evaluating localization.
For sparse scene reconstruction and camera calibration, we feed all the recorded (\textit{map} and \textit{query}) images into the COLMAP SfM pipeline \cite{schoenberger2016sfm,schoenberger2016mvs} to obtain high-quality camera calibrations.
The obtained camera poses of the query images serve as ground-truth $\boldsymbol{\hat{R}}$ and $\boldsymbol{\hat{T}}$ for our evaluations.
All query images alongside their corresponding 3D points are then carefully removed from the obtained reconstructions to prepare the 3D map for localization.
Afterwards, we perform another bundle adjustment with fixed camera poses to optimize the remaining 3D points given only the map images.
These steps are to reconstruct accurate ground-truth poses for the query images, and to also ensure a realistic 3D map for localization, in which we are only given the map images.
Across the datasets, we captured 375 single-image and 402 multi-image queries.

\customparagraph{Protocol.}
To establish 2D--3D correspondences, we use indirect matching of SIFT features at the default settings of the SfM pipeline \cite{schoenberger2016sfm,schoenberger2016mvs}.
In the single-image scenario, we treat each query image separately, while for the multi-image scenario, we group several consecutive images in the camera stream as one generalized camera.
When evaluating the multi-image case and pose estimation with known structure, we reconstruct the 3D points $\boldsymbol{\tilde{X}}$ and camera poses $\boldsymbol{P}_c$ using SfM \cite{schoenberger2016sfm,schoenberger2016mvs} from only the query images.
For a fair comparison, all methods use exactly the same 2D--3D correspondences, thresholds, and RANSAC implementation~\cite{fischler1981random}.
See supplementary material for more details.

\customparagraph{Metrics.}
In our evaluation, we compute the rotational error as $\Delta R = \arccos{ \tfrac{ \text{Tr}(\boldsymbol{R}^T \boldsymbol{\hat{R}}) - 1 }{2} }$ and the translational error as $\Delta T = \Vert \boldsymbol{R}^T \boldsymbol{T} - \boldsymbol{\hat{R}}^T \boldsymbol{\hat{T}} \Vert_2$.
We also report the average point-to-point and point-to-line (\cf Eq.~\eqref{eq:point-to-point-refinement} and Eq.~\eqref{eq:point-to-line-refinement}) reprojection errors \wrt to the obtained pose estimate.

\customparagraph{Methods.}
We compare the results of our proposed 8 privacy preserving to the corresponding 8 variants of traditional pose estimators, see Table~\ref{table:methods}.
The initial pose estimates of all methods are computed using standard RANSAC and a minimal solver for the geometric constraints.
We also compare the results of a non-linear refinement (suffix \textit{+ref}) of the initial pose using a Levenberg-Marquardt optimization of Eqs.~\eqref{eq:point-to-point-refinement}~and~\eqref{eq:point-to-line-refinement} based on the inliers from RANSAC.

%
%
%
\begin{table*}[t]
\vspace*{-2ex}
    \scriptsize
    \begin{tabularx}{1.0\textwidth}{L|L L|L}
        \toprule
        \multicolumn{2}{c}{\small\textbf{\textsc{Point to Point}} (Traditional)} &
        \multicolumn{2}{c}{\small\textbf{\textsc{Point to Line}} (Privacy Preserving)}  \\

        \cmidrule(rr){1-2} \cmidrule(ll){3-4}

        \hspace*{2.2ex} \textbf{\textit{p}3P}~\hspace*{4.5ex}~\hspace*{1ex}~\textbf{i:}~31  \textbf{r:}~64 \textbf{s:}~2.05 \textbf{t:}~3.54	               &
        \hspace*{2.1ex} \textbf{\textit{p}2P+u}~\hspace*{4.55ex}~\hspace*{1ex}~\textbf{i:}~15  \textbf{r:}~64 \textbf{s:}~2    \textbf{t:}~3.21              &
        \hspace*{2.2ex} \textbf{\textit{p}6L}~\hspace*{3.2ex}~\hspace*{1ex}~\textbf{i:}~158 \textbf{r:}~69 \textbf{s:}~64   \hspace*{0.6ex}\textbf{t:}~1002  &
        \hspace*{1.7ex} \textbf{\textit{p}4L+u}~\hspace*{4.4ex}~\hspace*{1ex}~\textbf{i:}~40~\textbf{r:}~70~\textbf{s:}~4~\textbf{t:}~5.27 \\

        \cmidrule(rr){1-2} \cmidrule(ll){3-4}

        \textbf{m-\textit{p}3P~\hspace*{4.3ex}}~\hspace*{1.15ex}~\textbf{i:}~38 \textbf{r:}~64 \textbf{s:}~1.70 \textbf{t:}~3.81            &
        \textbf{m-\textit{p}2P+u}~\hspace*{4.3ex}~\hspace*{1.3ex}~\textbf{i:}~11 \textbf{r:}~65 \textbf{s:}~2 \textbf{t:}~3.16             &
        \textbf{m-\textit{p}6L}~\hspace*{3.4ex}~\hspace*{1ex}~\textbf{i:}~\hspace*{1.2ex}64 \textbf{r:}~72 \textbf{s:}~64 \hspace*{0.6ex}\textbf{t:}~691               &
        \hspace*{-0.5ex}\textbf{m-\textit{p}4L+u}~\hspace*{4.5ex}~\hspace*{1ex}~\textbf{i:}~23~\textbf{r:}~72~\textbf{s:}~4~\textbf{t:}~3.61 \\

        \cmidrule(rr){1-2} \cmidrule(ll){3-4}
        \textbf{m-P3P+$\lambda$}~\hspace*{1.6ex}~\hspace*{1ex}~\textbf{i:}~23 \textbf{r:}~62 \textbf{s:}~1   \hspace*{2.2ex} \textbf{t:} 1.82   &
        \textbf{m-P2P+$\lambda$+u}~\hspace*{1.6ex}~\hspace*{1ex}~\textbf{i:}~12 \textbf{r:}~62 \textbf{s:}~1    \textbf{t:} 0.97   &
        \textbf{m-P4L+$\lambda$}~\hspace*{0.35ex}~\hspace*{1ex}~\textbf{i:}~\hspace*{1.2ex}27 \textbf{r:}~68 \textbf{s:}~1.5 \textbf{t:} 26.3   &
        \hspace*{-0.5ex}\textbf{m-P3L+$\lambda$+u}~\hspace*{1.5ex}~\hspace*{1ex}~\textbf{i:}~24 \textbf{r:}~68 \textbf{s:}~1    \textbf{t:} 4.07 \\

        \cmidrule(rr){1-2} \cmidrule(ll){3-4}
        \textbf{m-P3P+$\lambda$+s}~\hspace*{1ex}~\textbf{i:}~24 \textbf{r:}~62 \textbf{s:}~1    \hspace*{2.2ex} \textbf{t:}~4.19                 &
        \textbf{m-P2P+$\lambda$+u+s}~\hspace*{1ex}~\textbf{i:}~12 \textbf{r:}~62 \textbf{s:}~1    \textbf{t:}~2.37               &
        \textbf{m-P3L+$\lambda$+s}~\hspace*{-0.2ex}~\textbf{i:}~\hspace*{1.2ex}18 \textbf{r:}~68 \textbf{s:}~2.1 \textbf{t:}~2.3                 &
        \hspace*{-0.5ex}\textbf{m-P2L+$\lambda$+u+s}~\hspace*{1ex}~\textbf{i:}~\hspace*{1ex}9  \textbf{r:}~68 \textbf{s:}~2    \textbf{t:}~2.10 \\

        \midrule
    \end{tabularx}
    \scriptsize{\hspace*{1.2ex} Notation: \textbf{i:} mean number of iterations, \textbf{r:} inlier ratio [\%], \textbf{s:} mean number of solutions, \textbf{t:} minimal solver time [ms].}
    \label{table:RansacStatistics}
\end{table*}

%
%
%
%
\begin{table*}[t]
	\vspace{-4pt}
    \scriptsize
    \begin{tabularx}{1.0\textwidth}{L|L Y|Y}
        \cmidrule(rr){1-2} \cmidrule(ll){3-4}
        \hspace*{2ex} \textbf{\textit{p}3P}~\hspace*{4.7ex}~\hspace*{5ex}~1.84 / 1.42  &
        \hspace*{2.1ex} \textbf{\textit{p}2P+u}~\hspace*{4.6ex}~\hspace*{5ex}~1.88 / 1.43 &
        \hspace*{2.4ex} \textbf{\textit{p}6L}~\hspace*{4.6ex}~\hspace*{1ex}~1.55 (4.20) / 1.10 (3.24) &
        \hspace*{2.4ex}\textbf{\textit{p}4L+u}~\hspace*{4.4ex}~1.45~(3.62)~/~1.08~(3.11) \\

        \cmidrule(rr){1-2} \cmidrule(ll){3-4}
        \textbf{m-\textit{p}3P~\hspace*{4.5ex}}~\hspace*{5ex}~2.06 / 1.51 &
        \textbf{m-\textit{p}2P+u}~\hspace*{4.6ex}~\hspace*{5ex}~1.88 / 1.51 &
        \textbf{m-\textit{p}6L}~\hspace*{4.3ex}~\hspace*{1.4ex}~1.56 (4.23) / 1.13 (3.29) &
        \hspace*{-0.5ex}\textbf{m-\textit{p}4L+u}~\hspace*{4.5ex}~1.46~(3.90)~/~1.12~(3.17) \\
        \cmidrule(rr){1-2} \cmidrule(ll){3-4}

        \textbf{m-P3P+$\lambda$}~\hspace*{1.6ex}~\hspace*{5ex}~1.71 / 1.42 &
        \textbf{m-P2P+$\lambda$+u}~\hspace*{1.6ex}~\hspace*{5ex}~1.62 / 1.42 &
        \textbf{m-P4L+$\lambda$}~\hspace*{1.6ex}~\hspace*{1ex}~1.39 (4.20) / 1.08 (3.52) &
        \hspace*{-0.5ex}\textbf{m-P3L+$\lambda$+u}~\hspace*{1.6ex}~1.62~(4.92)~/~1.17~(3.73) \\
        \cmidrule(rr){1-2} \cmidrule(ll){3-4}

        \textbf{m-P3P+$\lambda$+s}~\hspace*{5ex}~1.72 / 1.43 &
        \textbf{m-P2P+$\lambda$+u+s}~\hspace*{5ex}~1.63 / 1.41 &
        \textbf{m-P3L+$\lambda$+s}~\hspace*{1ex}~1.47 (4.29) / 1.13 (3.48) &
        \hspace*{-0.5ex}\textbf{m-P2L+$\lambda$+u+s}~1.31~(4.06)~/~1.07~(3.52) \\
        \bottomrule
    \end{tabularx}
    \caption{\textbf{Quantitative Results.} RANSAC statistics (top) and reprojection errors in pixels (bottom) for \textit{initial / refined} results in traditional (Point to Point, Eq.~\eqref{eq:point-to-point-refinement}) and privacy preserving setting (Point to Line, Eq.~\eqref{eq:point-to-line-refinement}; in brackets also Point to Point, if we had the secret points).}
    \label{table:pixelMetric}
\end{table*}

\subsection{Results}

\vspace*{-0.5ex}
\customparagraph{Accuracy and Recall.}
The accuracy/recall curves are presented in Fig.~\ref{fig:errors_plots}, and reprojection errors in Table~\ref{table:pixelMetric}.
As expected, the traditional approaches achieve better accuracy/recall, because their solutions leverage two constraints for pose estimation.
Surprisingly, even though ours uses only a single geometric constraint, it comes very close to the results achieved by the traditional approach.
Moreover, incorporating known information about structure, gravity, and scale leads to an additional improvement of the results for all methods.

\vspace*{-0.5ex}
\customparagraph{Runtime.}
Table~\ref{table:pixelMetric} reports the mean number of required RANSAC iterations, inlier ratio, generated solutions in the minimal solver, and time required to estimate a single minimal solution.
The results show that, while our method is slower than the conventional approach, it provides runtimes that are suitable for practical real-time applications. Especially, the specialized solvers with known structure and gravity achieve competitive runtime.
We used the same RANSAC threshold for all the methods, but in practice this threshold could be chosen smaller for the privacy preserving methods, since the point-to-line is always smaller than point-to-point reprojection errors. This, together with the possibility to more easily include some additional outliers along the line due to mistakes in feature matching, leads to slightly higher inlier ratios for our method, see Table~\ref{table:pixelMetric}.

\begin{figure}[t!]
\centering
\includegraphics[width=\columnwidth]{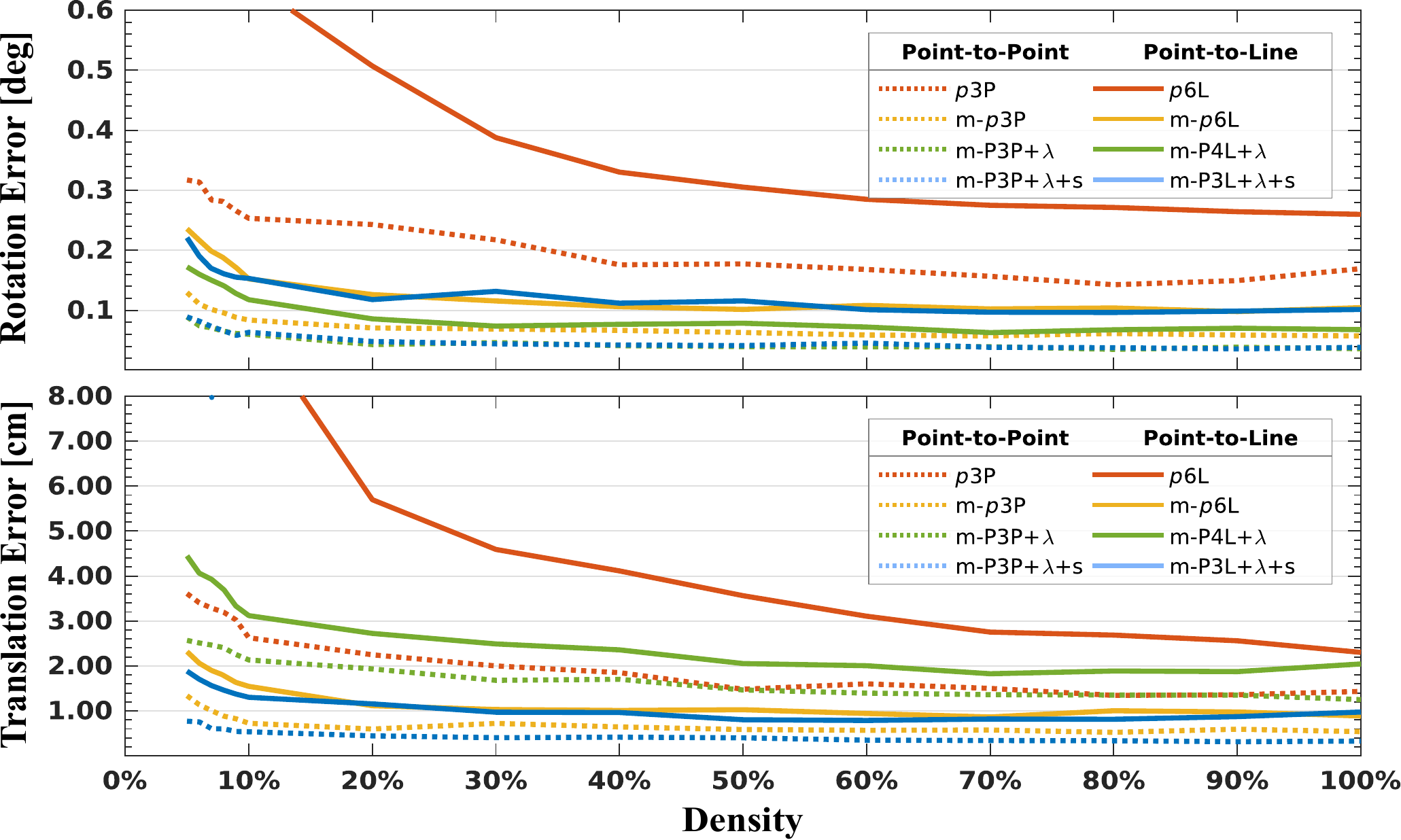} \\
\caption{\textbf{Point Cloud Density.} Rotation and translation error with varying point cloud density by uniformly dropping a percentage of points/lines at random from the map.}
\label{fig:sparsification}
\end{figure}

\begin{figure}[t!]
   \centering
   \includegraphics[width=\columnwidth]{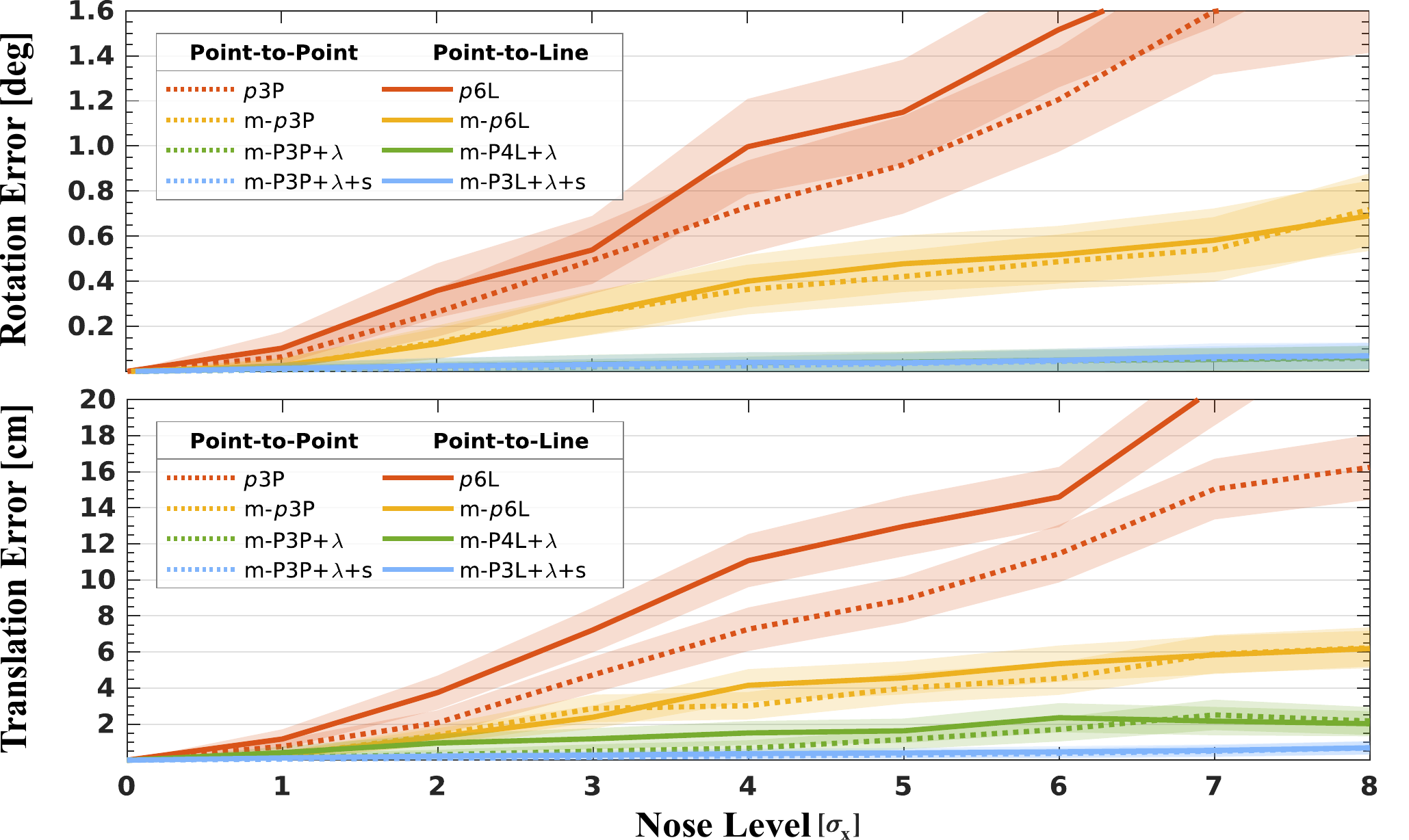} \\
   \caption{\textbf{Measurement Noise Sensitivity.} Rotation and translation errors for varying Gaussian noise level on the measurements.}
   \label{fig:synthetic_data}
\end{figure}

\vspace*{-0.7ex}
\customparagraph{Robustness.}
We study robustness with respect to point cloud density and image noise.
In Fig.~\ref{fig:sparsification}, we demonstrate reliable pose estimation even when only retaining every $20^\text{th}$ point of an already sparse SfM point cloud.
Fig.~\ref{fig:synthetic_data} shows similar behavior for ours and the conventional approach under varying noise $\boldsymbol{\sigma_x}$ on the image observations.

%
%




\section{Discussion}


Let us now discuss to what extent the privacy risks have been addressed, and highlight directions for future work.

\customparagraph{What is revealed during localization?}
When images are successfully localized within a scene, the inliers of the pose estimate reveal the secret 3D points through intersection of the camera rays with the corresponding 3D lines.
On first sight, this might seem like a privacy issue, but in practice only objects visible in the image are revealed, while the rest of the map or any confidential objects remain secret.

\customparagraph{Permanent Line Cloud Transformation.}
The lifting transformation must be performed only once and becomes permanent for a scene; otherwise, an adversary that retains multiple copies of line clouds generated by different lifting transformations
can easily recover the secret 3D points by intersecting corresponding 3D lines.

\customparagraph{Compactness of Representation.}
A more compact representation than Pl\"ucker lines in Eq.~\eqref{eq:plucker-line} would be to chose a finite set of line directions; \eg, 256 to fit within a byte, and encode the position of the line as the intersection with the plane through the origin and orthogonal to the direction, this reduces memory usage to 2 floats and 1 byte, \ie, even less than the 3 floats to encode a 3D point.  
%

\customparagraph{Privacy Attack on Line Clouds.}
Recovering the location of a single 3D point from its lifted 3D line representation is an ill-posed inversion problem, see Eq.~\eqref{eq:plucker-line}.
However, by analyzing the density of the 3D line cloud, one could~potentially recover information about the scene structure.
While 3D line clouds appear effective at making the underlying scene geometry incomprehensible,
it really depends on the sampling density of the 3D points in the scene (see suppl.~material).
In practice, we believe that our method is generally quite robust to such attacks, since image-based localization typically uses sparse SfM point clouds.
Besides, a sparsification in Fig~\ref{fig:sparsification} of the 3D line cloud is an effective defense mechanism.
Nevertheless, a more thorough theoretical analysis
is an interesting avenue of future research.

\vspace*{-0.8ex}
\section{Conclusion}
\label{sec:conclusion}
This paper introduced a new research direction called \textit{privacy preserving image-based localization}.
With this work, we are the first to address potential privacy concerns associated with the persistent storage of 3D point cloud models, as required by a wide range of applications in AR and robotics.
Our proposed idea of using confidential 3D line cloud maps conceals the geometry of the scene, while maintaining the ability to perform robust image-based localization based on the standard feature matching paradigm.
There are numerous directions for future work and we encourage the community to investigate this problem.

{\small
\bibliographystyle{ieee}
\bibliography{abbreviation_short,bibliography}
}
\end{document}